\documentclass[runningheads]{llncs}
\usepackage{graphicx}
\usepackage{subcaption}
\usepackage{caption}
\usepackage{comment}
\usepackage{amssymb}

\begin{document}
\title{A Realistic Collimated X-Ray Image Simulation Pipeline}
%
%
\author{Benjamin El-Zein\inst{1} \and
Dominik Eckert\inst{2} \and
Thomas Weber\inst{2} \and
Maximilian Rohleder\inst{1} \and
Ludwig Ritschl\inst{2} \and
Steffen Kappler\inst{2} \and
Andreas Maier\inst{1}}
\authorrunning{B. El-Zein et al.}
%
\institute{Pattern Recognition Lab, Friedrich-Alexander University, Erlangen-Nuremberg, Germany \and
Siemens Healthineers, Forchheim, Germany\\
\email{benjamin.el-zein@fau.de}}
\maketitle              
\begin{abstract}
Collimator detection remains a challenging task in X-ray systems with unreliable or non-available information about the detectors position relative to the source.
This paper presents a physically motivated image processing pipeline for simulating the characteristics of collimator shadows in X-ray images. 
By generating randomized labels for collimator shapes and locations, incorporating scattered radiation simulation, and including Poisson noise, the pipeline enables the expansion of limited datasets for training deep neural networks.
We validate the proposed pipeline by a qualitative and quantitative comparison against real collimator shadows. Furthermore, it is demonstrated that utilizing simulated data within our deep learning framework not only serves as a suitable substitute for actual collimators but also enhances the generalization performance when applied to real-world data.

\end{abstract}
\section{Introduction}
In digital radiography, the detection of collimator-covered areas is essential to present diagnostically relevant regions to radiologists. Geometric alignment algorithms, as described in \cite{luckner2018estimation}, can be employed in X-ray systems with known extrinsic projection parameters. However, despite their availability, these often suffer from inaccuracies sabotaging effectiveness in practice. Due to the inherent geometrical variability in conventional X-ray systems, particularly with mobile flat panel detectors, precise information of the relative position to the detector is unavailable. Moreover, imprecise collimator movement further complicates the detection process, necessitating analysis within image domain. Contrary to a simplistic threshold-based approach, the identification of relevant areas is challenging due to the presence of physical effects like edge-blurring, noise, and scattered radiation. Even human visual perception faces difficulties due to these complexities, as depicted in Fig. 1.

\begin{figure}
    \centering
    \begin{subfigure}[t]{0.24\textwidth}
        \centering
        \includegraphics[height=65pt]{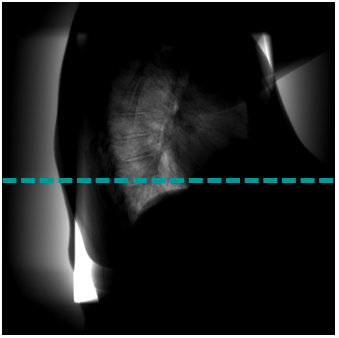}
        \caption{Full contrast}
    \end{subfigure}
    \hfill
    \begin{subfigure}[t]{0.24\textwidth}
        \centering
        \includegraphics[height=65pt]{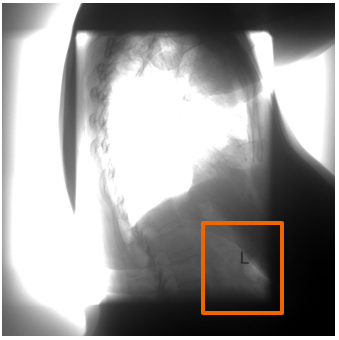}
        \caption{Contrast adjusted}
    \end{subfigure}
    \hfill
    \begin{subfigure}[t]{0.24\textwidth}
        \centering
        \includegraphics[height=65pt]{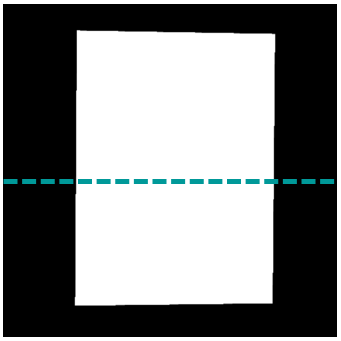}
        \caption{Collimator mask}
    \end{subfigure}
    \hfill
    \begin{subfigure}[t]{0.25\textwidth}
        \centering
        \includegraphics[height=65pt]{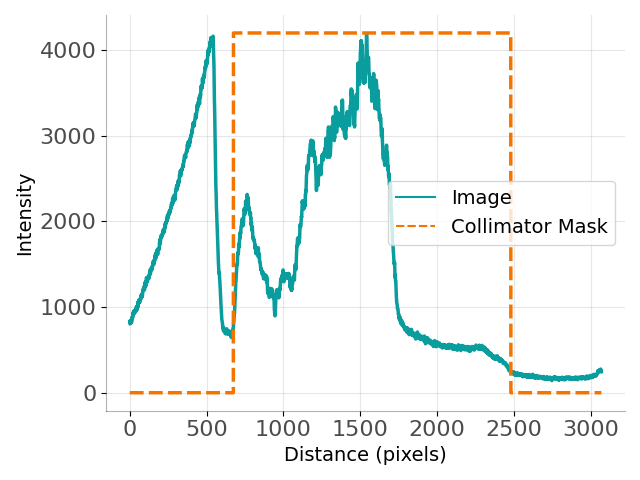}
        \caption{Lineplot}
    \end{subfigure}
    \caption{Illustrative case for collimator detection depicted in two contrast settings. (a) Contrast adjusted to full image. (b) Contrast adjusted to the orange box. The collimated area (c) is shown as a binary mask. In (d), the intensity profile along the dashed line is compared to the collimated area to visualize the complexity of image-based collimator detection.}
\end{figure}

Deep neural networks (DNNs) show promise for collimator detection, but the limited availability of pre-processed raw data poses a challenge for training robust networks in medical applications. So far, machine learning approaches for collimator detection have not significantly outperformed classic analytical methods in the literature. For instance, comparing the plane detection Hough transform proposed by Kawashita et al. \cite{kawashita2003collimation} with Mao et al.'s \cite{mao2014multi} approach that combines random forest learning with a landmark detector in a multi-view learning approach, both methods demonstrate similar performance on unseen data. According to Mao et al. \cite{mao2014multi}, each classifier was trained using only 200 training images.

To enhance the performance of machine learning algorithms, it is reasonable to assume that the implementation of robust data augmentation techniques is beneficial. These techniques aim to increase the quantity and variety of datasets. In this context, suitable augmentation techniques can be categorized into deep learning-based methods, such as generative adversarial networks (GANs) \cite{10.1145/3422622}, and physically motivated approaches. Although GANs have shown promising potential for post-processed X-ray image augmentation (without collimators) in studies like Bowles et al. \cite{bowles2018gan}, Madani et al. \cite{madani2018chest}, Kora et al. \cite{kora2020evaluation}, and Ng et al. \cite{ng2023generative}, they require sophisticated techniques and lack comprehensibility when aiming to serve as reliable training data.

Unlike this concept, physically motivated approaches offer a robust alternative for augmentation. These methods leverage an understanding of the underlying physics involved in imaging processes. By incorporating physical principles, these approaches ensure reproducibility and reliability, as demonstrated by Eckert et al. \cite{eckert2020deep} and Xu et al. \cite{xu2022physics}.

In this paper a physically motivated image processing pipeline is presented that simulates the characteristics of real collimators enabling the expansion of limited datasets of X-ray images without collimators. The data augmentation method enables the generation of unlimited pre-processed image data e.g. for training DNNs.

\section{Methods}

\subsection{Randomized Collimator Simulation Pipeline}
The process of simulating collimator shadows involves three distinct stages. In the first stage random labels are generated to define the shape and position of the collimated area. The second stage introduces scattered radiation, whereas the final stage adds a simulation of noise.

\subsubsection{Binary Mask Sampling Strategy:}
The first part of the pipeline generates a binary mask of the same shape as the X-ray image to be investigated. Thus, all image pixels are classified according to their property of lying inside or outside the assumed collimator shadow. To determine random position, shape and size of the collimator, a centroid location along with width and height are sampled from a truncated normal distribution, yielding a rectangle as shown in Fig.2a. Due to practical constraints in clinical settings, e.g. with bedridden patients, the resulting images may exhibit rotations or distortions that deviate from the desired orientation. To accommodate these cases, a randomized rotation and distortion transformation are applied to the binary image's rectangle, as shown in Fig.2b and Fig.2c.

\begin{figure}
    \centering
    \begin{subfigure}[h]{0.27\textwidth}
        \centering
        \includegraphics[width=\textwidth]{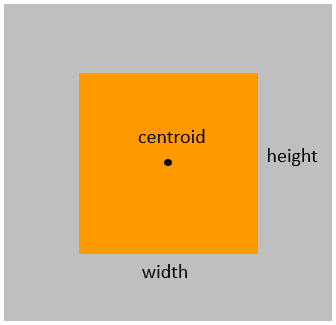}
        \caption{Binary mask}
    \end{subfigure}
    \hfill
    \begin{subfigure}[h]{0.27\textwidth}
        \centering
        \includegraphics[width=\textwidth]{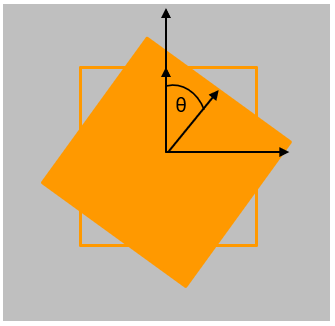}
        \caption{Rotation}
    \end{subfigure}
    \hfill
    \begin{subfigure}[h]{0.27\textwidth}
        \centering
        \includegraphics[width=\textwidth]{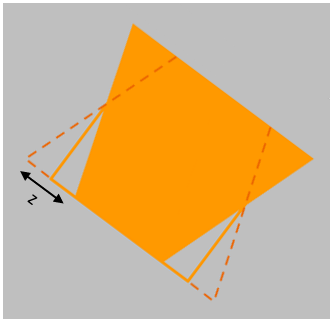}
        \caption{Distortion}
    \end{subfigure}
    \caption{Example of a rectangular binary mask being transformed afterwards by rotation and shape distortion to cover the range of essential deviations in clinical practice.}
\end{figure}

\subsubsection{Collimator Physics:}
To account for collimator attenuation, the binary mask is adjusted by assigning its zeros to a damping factor, resulting in the mask $M_d$. The non-infinitesimal size of the focal spot causes blurring of the collimator edges at the detector as shown in Fig. \ref{fig_focal_spot}. As the intensity profile of the radiated X-rays is Gaussian distributed in space, this effect can be approximated by convolving the mask $M_d$ with a Gaussian kernel $G_b$. Hence, the damping operation can be applied to the input image $I_{input}$ as follows: 

\begin{equation}
\label{eq_focal_spot}
    L(I_{input}) = (M_{d} \ast G_b) \cdot  I_{input}
\end{equation}

\subsubsection{Scattered Radation Simulation:}
\begin{figure}
    \centering
    \hspace{0.7cm}
    \begin{subfigure}[b]{0.4\textwidth}
        \centering
        \includegraphics[width=\textwidth]{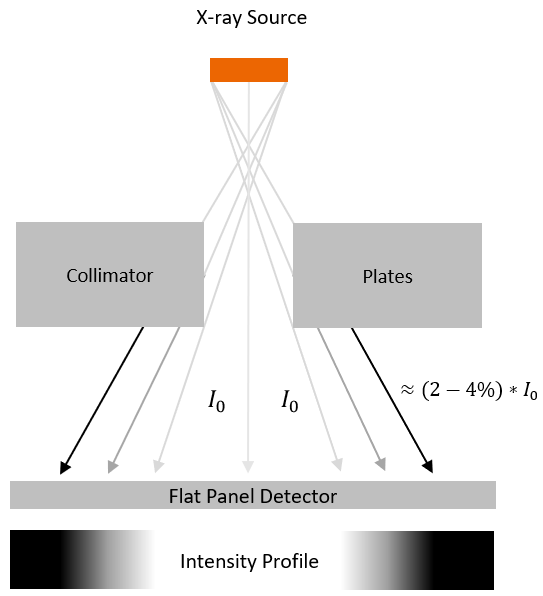}
        \caption{Focal spot characteristic}
        \label{fig_focal_spot}
    \end{subfigure}
    \hfill
    \begin{subfigure}[b]{0.4\textwidth}
        \centering
        \includegraphics[width=\textwidth]{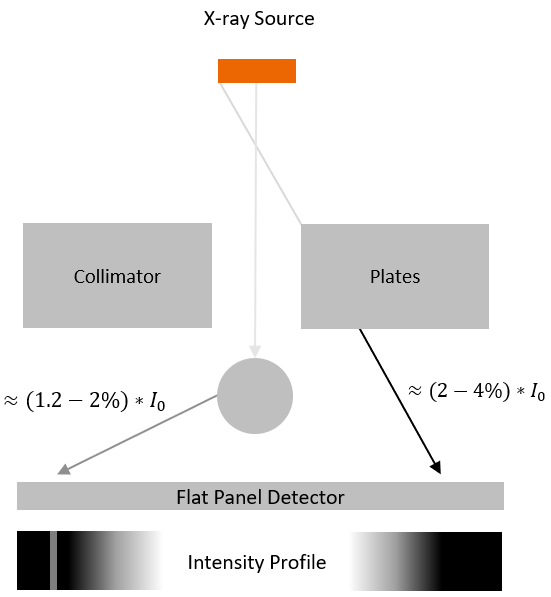}
        \caption{Influence of scattered radiation}
        \label{fig_scatter}
    \end{subfigure}
    \hspace{0.7cm}
    \caption{Physical properties to be considered when modeling collimators in a basic X-ray system. Edge-blurring introduced by the focal spot characteristic not being an ideal point, as well as increasing intensities within the collimated region due to photons that get scattered by the Compton effect.}
\end{figure}

Incoherent scattering describes the process of an high energetic photon colliding with matter, resulting in a deflection from the initial pathway as shown in Fig.\ref{fig_scatter} \cite{krieger2007grundlagen}. Due to this fact it predominantly affects X-ray imaging depending on the matter the photon interacts with. A collimator influences the number of photons that pass through the patient, altering the scatter characteristics. This change must be accounted for in the simulation. According to comprehensive Monte Carlo simulation studies \cite{sisniega2013monte}, it was shown that intensities created by scattered photons are in a range from 1.2\%-2\% of the primary intensity, e.g. for thorax images of c-arm systems without anti scatter grids. In relation to the dampened intensities by the collimator, contributing 2\%-4\% of the primary intensity based on empirical analysis, scatter has a significant influence.

\subsubsection{Scatter Estimation:}
The intended pipeline requires a methodology capable of modeling the distribution of scattered photons in a collimated X-ray image. Ohnesorge et al. \cite{ohnesorge1999efficient} present a convolution kernel based scatter estimation that can be utilized for our application. At first, they define a scatter potential $S_{p}$ which is defined as follows: 

\begin{equation}
    S_{p}(I|I_{0}) = c \cdot \left(\frac{I}{I_{0}}\right)^\alpha \cdot \ln\left(\frac{I_{0}}{I}\right)^\beta
\end{equation}

The input image $ I $ and the primary intensity $I_0$ are modified by three hyper parameters $\alpha$, $\beta$ and $c$.

Finally, the estimated scatter $S_{e}$ is obtained by convolving the scatter potential with a Gaussian kernel $G_{s} $ as demonstrated in this equation:

\begin{equation}
    S_{e}(I) = \left(S_{p}(I|I_{0}\right) \ast G_{s}) \cdot I_{0}
\end{equation}

\subsubsection{Scatter Correction:} 
Given the presented framework, this part of the pipeline follows a two-step process. First, the scatter present in $I_{input}$ is removed, as it does not match the scatter of a real collimator specified by $L(I_{input})$. To achieve this, we employ the scatter estimation method proposed by Ohnesorge et al., as depicted by the following equation, in order to obtain a scatter-free image, $I_{sc}$.

\begin{equation}
    I_{sc} = I_{input} - S_{e}(I_{input})
\end{equation}

In the second step, the collimated image $I_s$ with the corresponding scatter is simulated. Eq. \ref{eq_focal_spot} is applied to collimate the scatter free image. The corresponding scatter map is generated and added to the image as following: 

\begin{equation}
    I_{s} = L(I_{sc}) + S_{e}(L(I_{sc}))
\end{equation}

\subsubsection{Poisson Noise Simulation:}
Due to the quantum properties of light, photons exhibit random arrival times. Hence, there exists a level of uncertainty regarding the received signal. The probability for z photons arriving at one pixel at the detector can be modeled by the Poisson distribution, which is defined as
\begin{equation}
    P(z | \lambda) = \frac{\lambda^ze^{-z}}{z!} \, .
\end{equation}
It is dependent on the parameter $ \lambda $ that represents the average rate at which an event occurs, thus the mean arrival rate of the photons.

The damping of real collimators by a factor $ \alpha=[0,1]$ causes a reduction of the mean photon arrival rate  $\lambda $ to $ \alpha \lambda$ and hence, an increase in noise in the image.
Since for a Poisson distribution $ \lambda = \sigma = \mu $ holds true, the altered SNR is defined with a new mean $ \mu_n$ and variance $ \sigma_n$ respectively as
\begin{equation}
    \textrm{SNR} = \frac{\mu_n}{\sigma_n} = \frac{\alpha \lambda}{\sqrt{\alpha \lambda}} = \sqrt{\alpha} \, .
\end{equation}
Therefore, besides scaling the intensities in the region of the simulated collimator, the noise level has to be increased to account for the increased uncertainty in the number of arrived photons.
So far, applying our collimator mask did change $ \mu $ to $ \alpha \mu $ and  $ \sigma $ to $ \alpha \sigma $, remaining the SNR unchanged. To compensate for this, we add a normal distribution $ N(0, \sigma_x)$ to the signal, to get the right SNR \cite{eckert2022deep}\cite{eckert2020deep}:
\begin{equation}
    \sqrt{\alpha}
    = \frac{\alpha \mu}{\sqrt{\sigma_{x}^2+ \alpha^2 \lambda}}
\end{equation}
Rearranging the equation yields $ \sigma_x = \sqrt{\lambda \cdot (1 - \alpha)}$.

\subsection{Experiments}
\subsubsection{Real vs. Simulated:}
For the purpose of evaluation, the pipeline's output is examined by acquiring X-ray images of an anthropomorphic thorax phantom. The acquisitions include both open field and a collimated image, approximately to the lungs.
We use our pipeline to simulate a matching collimator on the open field X-ray image and compare that image to the physically collimated image. To perform a detailed analysis, various image patches are extracted and quantitatively analyzed.

\subsubsection{Application Case DNN:} Using the augmentation method described, we evaluate its effectiveness in simulating collimators within a simple deep learning framework. We generate samples and random labels on-the-fly for 1500 real in-house X-ray images with collimators being manually cropped out. This is called SimNet. In addition, we trained a second DDN, referred to as RealNet, with the uncropped images containing the real collimators and hand-labeled masks. This allows to inspect if it is possible to replace real collimated images with simulated images of our pipeline. 

Both DNNs are based on the DeepLabV3 architecture \cite{Chen2017RethinkingAC}, classifying pixels as collimated or non-collimated, utilizing the Dice metric as a loss function and the ADAM optimizer during 500 epochs of training. Evaluation is performed on three datasets calculating the Dice score: a subset of 80 randomly extracted training images, 30 challenging cases with dark attenuating line-shaped implants, and 20 images showing detector line artifacts. RealNet is evaluated with the real collimator version. Furthermore, we check SimNets performance on real and simulated versions of the test sets data in order to reveal the pipelines authenticity and generalization on authentic data.

\section{Results} 
\subsection{Framework Validation}
Fundamentally, the goal of the simulation pipeline is to generate real collimator intensity distributions. On the one hand, the Figures below show that the desired image impression can be achieved.

\begin{figure}[!htb]
    \centering
    \begin{subfigure}[b]{0.22\textwidth}
        \centering
        \includegraphics[width=\textwidth]{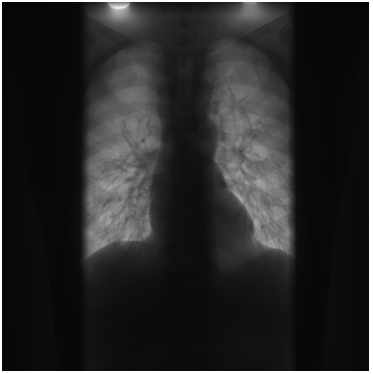} 
        \caption{Real: anatomy contrast}
    \end{subfigure}
    \hfill
    \begin{subfigure}[b]{0.22\textwidth}
        \centering
        \includegraphics[width=\textwidth]{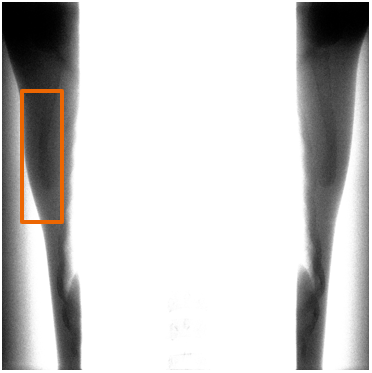}
        \caption{Real: collimator contrast}
    \end{subfigure}
    \hfill
    \begin{subfigure}[b]{0.22\textwidth}
        \centering
        \includegraphics[width=\textwidth]{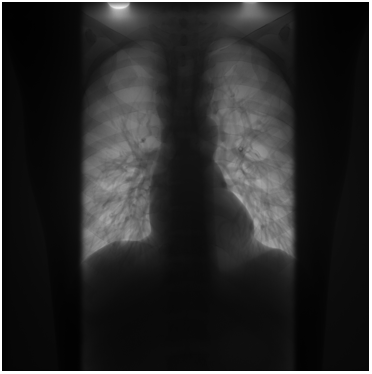}
        \caption{Simulated: anatomy contrast}
    \end{subfigure}
    \hfill
    \begin{subfigure}[b]{0.22\textwidth}
        \centering
        \includegraphics[width=\textwidth]{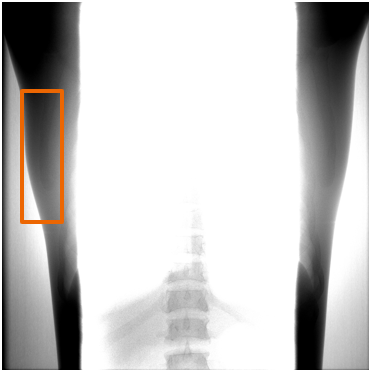}
        \caption{Simulated: collimator contrast}
    \end{subfigure}
    \caption{Comparison of a real collimated image with the ouput of the pipeline based on an open field image acquired by the same setup. Both of the images are shown in two different contrast ranges. Besides being in the complete value range, intensities are limited to the indicated box regions.}
\end{figure}

On the other hand, uncertainties arising from the inherent approximations made within the pipeline can be identified. The scatter's Gaussian distribution, attributable to the simulation methodology, is distinctly discernible and exhibits a discrepancy from the actual scatter behavior. To obtain a more detailed representation of this observation, we proceed with showing the intensity distribution along the dashed line indicated in Fig.6. Leveraging the understanding of the impact of a real collimator on images is achieved by showing the real collimator image together with the open field image in Fig.5a. Subsequently, Fig.5b presents the same for the real collimator and the simulated collimator, confirming high resemblance from qualitative point of view.

\begin{figure}
    \centering
    \begin{subfigure}[c]{0.49\textwidth}
        \centering
        \includegraphics[width=\textwidth]{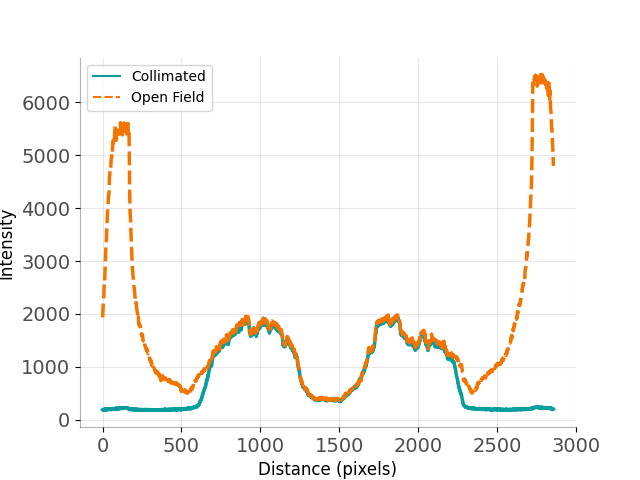}
        \caption{Open Field vs. Real Collimator}
    \end{subfigure}
    \begin{subfigure}[c]{0.49\textwidth}
        \centering
        \includegraphics[width=\textwidth]{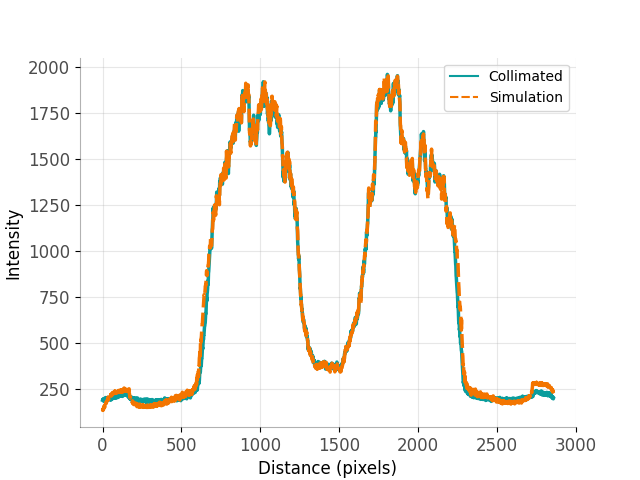}
        \caption{Real vs. Simulated Collimator}
    \end{subfigure}
    \caption{Line plots comparing real collimator damping on a real X-ray image as well as presenting the differences between the real and simulated collimator.}
\end{figure}

Furthermore, real and simulated collimator are quantitatively examined by the normalized mean-squared-error (nMSE), the structural similarity index (SSIM) and the Peak SNR (PSNR) of the image patches depicted in Fig.5. Table 1 shows that the non-collimated region specifically exhibits an almost identical characteristic. The patches within collimator region however demonstrate that slight diverging behaviour exists, but does not exceed the requirement of being very similar. In particular this is proven by still showing a very high score for the SSIM.

\begin{figure}[!htbp]
  \centering
  \begin{minipage}[t]{0.4\textwidth}
    \centering
    \includegraphics[height=20mm]{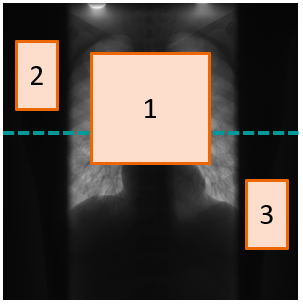}
    \caption{Orientational scheme showcasing the location of the defined image patches as well as the line for the intensity value distributions.}
    \label{fig:image}
  \end{minipage}
  \hfill
  \begin{minipage}[t]{0.49\textwidth}
    \centering
    \vspace{-17mm}
    \begin{tabular}{l|ccc}
    \hline
    \hline
    \textbf{Patch} & \textbf{nMSE} & \textbf{SSIM} & \textbf{PSNR} \\
    \hline
    \hline
    1 & 0.0001 & 0.9998 & 33.2286\,[dB] \\
    2 & 0.0676 & 0.9962 & 31.7112\,[dB]  \\
    3 & 0.0192 & 0.9997 & 32.8377\,[dB]  \\
    \hline
    \end{tabular}
\captionof{table}{Statistical Measures of Image Patches comparing real and simulated collimator images.}
  \end{minipage}
\end{figure} 

\subsection{Network Evaluation} The performance of the DNN based on pipeline augmentation (SimNet), as presented in Table 2, demonstrates that it achieves slightly better results on simulated data, which can be attributed to its training on such data. However, the network's ability to perform very well on real-world data validates the concept and affirms that the images processed by simulation maintain a high level of realism. We can further prove this by showing that SimNet even exceeds the performance of RealNet. Future research will focus on exploring network architectures possessing explicit constraints for collimator detection, thereby enabling them to distinguish edges more efficiently.

\newcommand*{\myalign}[2]{\multicolumn{1}{#1}{#2}}

\begin{table}
 \centering
    \centering
    \begin{tabular}{l|cc|c}
     & \multicolumn{2}{|c|}{\textbf{SimNet}} & \textbf{RealNet} \\
    \hline
    \hline
    \textbf{Test sets} & Real & Simulated  & Real \\
    \hline
    \hline
    General Test & $\mathbf{0.9718} \pm0.027$ & $\mathbf{0.9749}\pm0.041$ & $\mathbf{0.9641}\pm0.048$ \\
    Line Artifacts Test & $\mathbf{0.9778}\pm0.025$ & $\mathbf{0.9873}\pm0.014$ & $\mathbf{0.9652}\pm0.038$ \\
    Implants Test & $\mathbf{0.9494}\pm0.071$ & $\mathbf{0.9820}\pm0.027$ & $\mathbf{0.9780}\pm0.015$\\
    \hline
    \end{tabular}
        \vspace{0.3cm}
    \caption{RealNet vs. SimNet Dice score performance comparison.}
\end{table}

\section{Discussion}
The presented X-ray image processing pipeline effectively simulates the properties of real collimators by incorporating key physical effects such as scattered radiation and quantum noise, adapting them to the randomly generated labels that define the synthetic collimators shape and location. This approach enables the generation of an unlimited amount of collimator training data. The realism of the generated characteristics is both qualitatively and quantitatively validated. When applied to a dataset of real X-ray images within a DNN environment, the pipeline demonstrates a significantly close performance on real test collimators compared to the simulated ones. Furthermore, it outperforms the corresponding DNN based on training with real collimators and hence no augmentation, affirming the effectiveness of the proposed approach.

\vspace{0.5cm}
\noindent\textbf{Disclaimer:} The concepts and information presented in this paper are based on research and are not commercially available.
%
%
%
\bibliography{main}


\end{document}